\documentclass{article}

\PassOptionsToPackage{numbers}{natbib}
\usepackage[preprint]{neurips_2024}

\usepackage[utf8]{inputenc} 
\usepackage[T1]{fontenc}    
\usepackage{hyperref}       
\usepackage{url}            
\usepackage{booktabs}       
\usepackage{amsfonts}       
\usepackage{nicefrac}       
\usepackage{microtype}      
\usepackage{xcolor}         

\usepackage{graphicx}
\usepackage{algorithm}
\usepackage{algorithmic}
\usepackage{amsmath}

\usepackage{mwe}
\usepackage{caption}

\title{Curriculum-enhanced GroupDRO: Challenging the Norm of Avoiding Curriculum Learning in Subpopulation Shift Setups}

\author{%
  Antonio Barbalau \\
  Bitdefender, Bucharest, Romania\\
  University of Bucharest, Romania \\
  \texttt{abarbalau@bitdefender.com} \\
}

\begin{document}

\maketitle

\begin{abstract}

In subpopulation shift scenarios, a Curriculum Learning (CL) approach would only serve to imprint the model weights, early on, with the easily learnable spurious correlations featured. To the best of our knowledge, none of the current state-of-the-art subpopulation shift approaches employ any kind of curriculum. To overcome this, we design a CL approach aimed at initializing the model weights in an unbiased vantage point in the hypothesis space which sabotages easy convergence towards biased hypotheses during the final optimization based on the entirety of the available data. We hereby propose a Curriculum-enhanced Group Distributionally Robust Optimization (CeGDRO) approach, which prioritizes the hardest bias-confirming samples and the easiest bias-conflicting samples, leveraging GroupDRO to balance the initial discrepancy in terms of difficulty. We benchmark our proposed method against the most popular subpopulation shift datasets, showing an increase over the state-of-the-art results across all scenarios, up to $6.2\%$ on Waterbirds.
\end{abstract}

\section{Introduction}


Curriculum learning \cite{curriculum, survey_sov, survey_wang} is a widely adopted strategy in deep learning, serving to improve model generalization capabilities and convergence time, by means of incrementally increasing training difficulty, starting from the easiest samples, and gradually moving to harder ones as the learning progresses. However, as noted by Wang et al. \cite{survey_wang}, not all circumstances benefit from a curriculum learning approach; at least, not in its standard easiest-first design. In a subpopulation shift scenario \cite{change_is_hard}, the training data naturally presents itself with strong correlations between certain environments and certain classes. As an example, a model trained using the standard Empirical Risk Minimization (ERM) procedure on the Waterbirds \cite{waterbirds} benchmark will associate water backgrounds with the waterbird class and land backgrounds with the landbird class. These spurious cues are easy to learn and convenient to use as biases to short-circuit decisions, leaving the model vulnerable at test time, when the subpopulation of waterbirds on land backgrounds increases. In such a scenario, a standard curriculum approach would only serve to prioritize the easy biased samples, serving the spurious features as the first thing to be imprinted upon the model weights, deepening test-time inadequacy.

In fact, the latest state-of-the-art methods \cite{too_good_to_be_true,jtt,xrm,simplicity_bias,feed} on environment discovery for distributionally robust optimization rely precisely and explicitly on the observation that the easiest samples to learn are the ones which confirm the class-environment biases the most. Liu et al. (JTT) \cite{jtt} fit an ERM model on the training data and denote the samples classified correctly as bias-confirming, and the misclassified samples and bias-conflicting. Zare et al. (FEED) \cite{feed} remove the samples with the highest loss from the training pool after each epoch, in order to arrive at a subset containing only biased samples. We emphasize the pronounced simlilarity between FEED \cite{feed} and Self-Paced Curriculum Learning \cite{self-paced}, and the fact that it leads to biasing the model to the highest degree possible. 



\begin{minipage}{\linewidth}
  \centering
  \begin{minipage}{0.48\linewidth}
      \begin{figure}[H]
          \includegraphics[width=\linewidth]{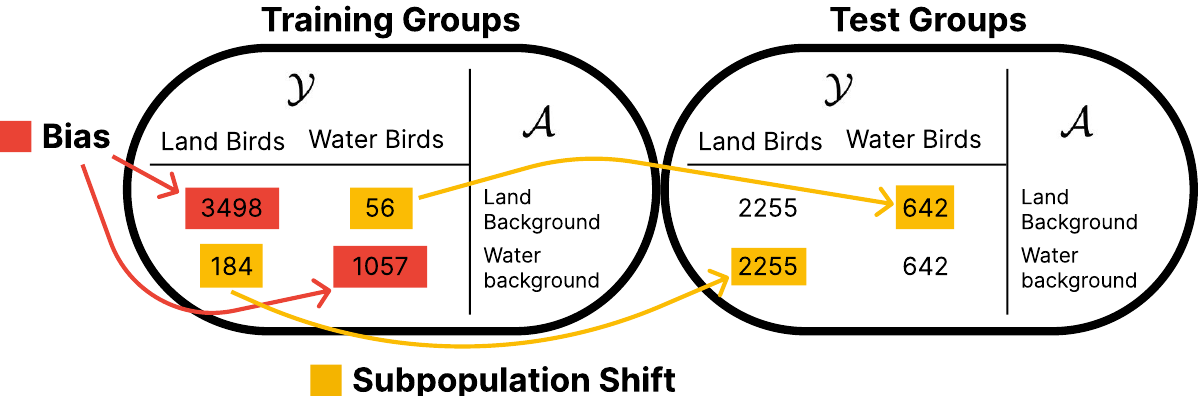}
          \caption{Sample distribution for Waterbirds, illustrating the subpopulation shift setup.}
          \label{cegdro:fig:subpop_shift}
      \end{figure}
  \end{minipage}
  \hspace{0.02\linewidth}
  \begin{minipage}{0.48\linewidth}
      \begin{figure}[H]
          \includegraphics[width=\linewidth]{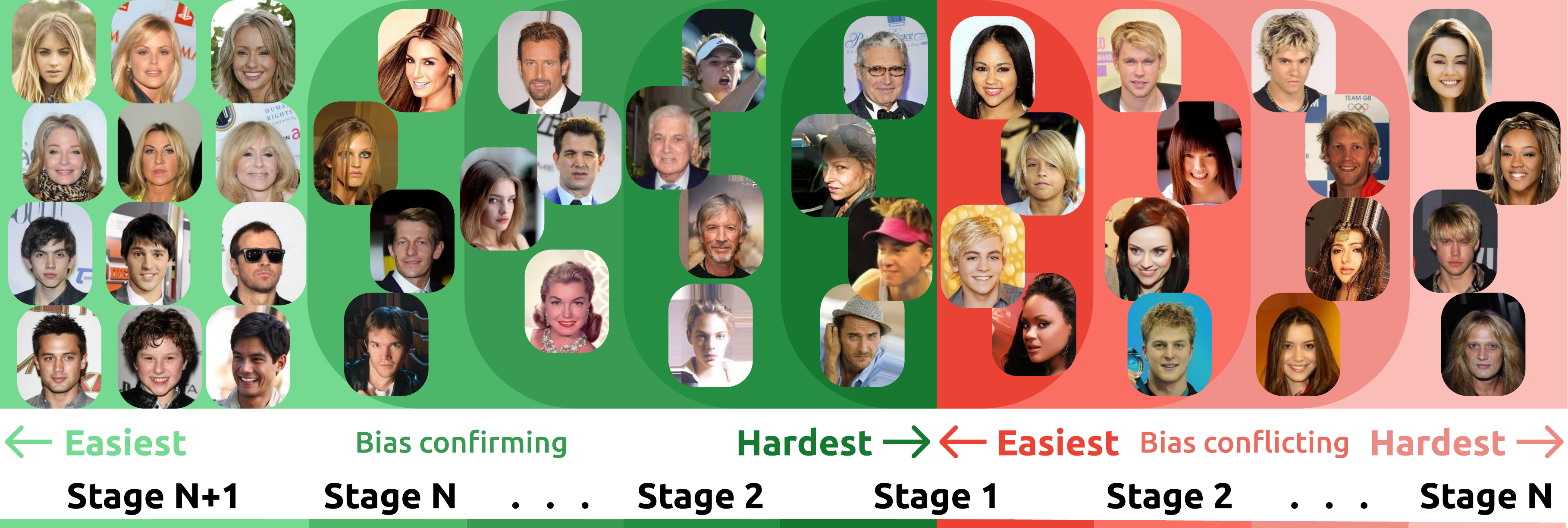}
          \caption{Illustration of our proposed Curriculum enhancement for GroupDRO on CelebA.}
          \label{cegdro:figure}
      \end{figure}
  \end{minipage}
\end{minipage}
\\

We aim to overcome this limitation and bring the benefits of curriculum learning into the subpopulation shift scenery. We hereby propose a novel general Curriculum Learning design, meant to enhance Group Distributionally Robust Optimization (GroupDRO) \cite{groupdro} for subpopulation shift setups. In its original form, GroupDRO adjusts the training loss across a set of given or discovered environments in order to address the class-environment imbalances featured in the training dataset. We further enhance this procedure by designing a training curriculum aimed at presetting the model weights, before the final GroupDRO training procedure, in an unbiased vantage point within the hypothesis space which sabotages easy convergence towards biased hypotheses. To this end, we leverage discovered environments to prioritize the easiest bias-conflicting samples and the hardest bias-confirming samples in equal amount during the curriculum procedure, balancing the loss discrepancy by means of the GroupDRO update rule, thus initializing the model weights without bias or with a very slight opposite bias, and delaying the association between the strongly correlated classes and environments for as long as possible. Our curriculum is illustrated in Figure \ref{cegdro:figure} for CelebA, where the bias consists in associating non-blonde hair with men and blonde hair with women. To the best of our knowledge, we are the first to propose a generic Curriculum Learning design for subpopulation shift setups. We evaluate our Curriculum-enhanced GroupDRO (CeGDRO) approach on the most popular benchmarks, achieving consistent improvements over the state-of-the art results, e.g.~up to $6.2\%$ on Waterbirds.

\section{Background}

In a subpopulation shift setup, the dataset features a set of spurious attributes $\mathcal{A}$, i.e. $\mathcal{D} = (\mathcal{X}, \mathcal{Y}, \mathcal{A})$, and each data sample $x_i \in \mathcal{X}$ has an associated spurious attribute $a_i \in \mathcal{A}$. Discrete subpopulation groups $g \in \mathcal{G}$ are formed based on some function $h: \mathcal{Y} \times \mathcal{A} \rightarrow \mathcal{G}$. We illustrate the sample distribution of Waterbirds \cite{waterbirds} over the training and test groups in Figure \ref{cegdro:fig:subpop_shift}, showcasing how the training dataset naturally emphasizes some groups more than others. In such a scenario, the average accuracy is not a good predictor of the real-world capabilities of the model, as it could just heavily rely on just predicting the spurious features. The end goal is thus to learn a set of weights which only accounts for the relevant features, yielding good performance across all environments by means of disregarding spurious features. Formally, given a loss function $\mathcal{L}$, the end goal, defined by Eq.~\eqref{cegdro:eq:objective}, is to optimize $M^*: \mathcal{X} \rightarrow \mathcal{Y}$ to minimize the worst group loss for a given target probability distribution:
\begin{equation}
    M^* = \arg \min_{M} \max_{g\in\mathcal{G}} \mathbb{E}_{p(x,y) \sim p_g} \mathcal{L}(y, M(x)).
    \label{cegdro:eq:objective}
\end{equation}
At test time, ground-truth groups are always used to evaluate the model. During training, current state-of-the-art approaches \cite{cobalt, eiil, jtt, xrm, feed} follow a two step procedure: firstly, an environment discovery method is employed to determine spurious attributes, and secondly, a distributionally robust optimization procedure is used to balance out these environments. In the first, step env-discovery methods can either determine semantic groups \cite{cobalt, b2t} or bias-confirming (easy) and bias-conflicting (hard) splits directly \cite{too_good_to_be_true, xrm, simplicity_bias, feed}. The current two main options for the second step are Invariant Risk Minimization (IRM) \cite{irm} and GroupDRO \cite{groupdro}. In the case of the latter, at each training step, for a given group $g$ and a sample $(x, y)$, a group weight $q_g$ is updated via the rule described in Eq.~\eqref{cegdro:eq:groupdro}. After this update is performed, the group weight is balanced with respect to that of other groups, that is,  $q_g \gets q_g / \sum_{g^\prime} q_{g^\prime}$. This serves to allocate high masses to high loss groups. 

\begin{equation}
    q_g \gets q_g \exp(\eta \mathcal{L}(y, M(x))), \; \text{where}\: \eta \: \text{is a hyperparameter.}
    \label{cegdro:eq:groupdro}
\end{equation}

Finally, these weights intervene to scale the gradient for samples in their corresponding group. In the context of our current proposal, we work to improve upon the GroupDRO post-environment discovery optimization protocol, proposed by Sagawa et al. \cite{groupdro}.

\section{Method}

\begin{algorithm}[tb]
\caption{Curriculum-enhanced GroupDRO}
\label{cegdro:alg:algorithm}
\textbf{Input}: $M$ - a deep learning model, $D$ - training dataset, $D_B \subset D$ - set of bias-confirming samples, $D_C \subset D$ - set of bias-conflicting samples.\\
\textbf{Parameters}: $R$ - rate of increase for the number of available samples, $E_s$ - number of epochs for each curriculum stage, $E_f$ - number of epochs for the final training stage.\\
\textbf{Output}: Trained model $M$.
\begin{algorithmic}[1] 

\STATE $M^\prime \gets \mbox{ERM}(M, D)$

\STATE $I \gets \mbox{reversed}(\mbox{argsort}(\mbox{Loss}(M^\prime, D_B)))$
\STATE $J \gets \mbox{argsort}(\mbox{Loss}(M^\prime, D_C))$

\STATE $P \gets R$
\WHILE{$P \le 1$}
    \STATE $N \gets \lfloor |D_C| * P) \rfloor$
    \STATE $S \gets \{D_{B_{I_1}}, D_{B_{I_2}}, \cdots, D_{B_{I_N}}\} \cup \{D_{C_{J_1}}, D_{C_{J_2}}, \cdots, D_{C_{J_N}}\}$
    \STATE \textbf{for} $e \gets 1$ to $E_s$ \textbf{do} $M \gets \mbox{GroupDRO}(M, S)$ \textbf{end for}

    \STATE $P \gets P + R$
\ENDWHILE

\STATE \textbf{for} $e \gets 1$ to $E_f$ \textbf{do} $M \gets \mbox{GroupDRO}(M, D)$ \textbf{end for}

\STATE \textbf{return} $M$
\end{algorithmic}
\end{algorithm}
Our Curriculum-enhanced GroupDRO method is formally described in Algorithm \ref{cegdro:alg:algorithm}. Prior to our procedure, we separate the training data, based on $\mathcal{G}$, in two main subsets: bias-confirming $\mathcal{D}_B \subset \mathcal{D}$ and bias-conflicting $\mathcal{D}_C \subset \mathcal{D}$, as described in Section \ref{cegdro:section:experiments}. We then, at line $1$, train an ERM model, $M^\prime$, featuring the same architecture as our final desired model, for a single epoch on the entire training set. The loss of $M^\prime$ with respect to $\mathcal{D}_B$ and $\mathcal{D}_C$ is used to sort out the samples from each subset. We store the sorted indices of samples from $\mathcal{D}_C$ in $J$ at line $3$, and the sorted indices of samples from $\mathcal{D}_B$, in reversed order, prioritizing the hardest ones, in $I$ at line $2$. At line $4$, we initialize the percentage of available samples $P$ with a hyperparameter $R$, which controls the rate of increase for the percentage of available samples at each stage during the procedure. The number of samples available at a given stage of the curriculum is $N = \lfloor |D_C| * P) \rfloor$, computed at line 6. We define the training subset $S$ for the current stage, by selecting the first $N$ samples from $\mathcal{D}_B$ and $\mathcal{D}_C$, in terms of the sorted indices from $I$ and $J$ respectively, at line $7$.

This selection ensures that the features of what will become, in the last stage of training, the worst represented groups, are readily available at the beginning. It further ensures that it is as hard as possible for the network to associate the biased features with their respective classes. Though this selection is likely to slightly imprint the model weights with the opposite bias, by means of providing an equal number of bias-conflicting and bias-confirming samples at each training stage, and by leveraging GroupDRO to balance the initial discrepancy in terms of difficulty, we aim to skim this likelihood as much as possible. At each stage, we train the model $M$ on the subset $S$ for $E_s$ epochs. We then increase $P$ by $R$ and continue the process while $P \leq 1$. When $P = 1$, the entirety of the bias-conflicting data is available for training, together with an equal amount of bias-confirming data. After the curriculum procedure reaches completion, we continue training the model for $E_f$ epochs on the entirety of $\mathcal{D}$, while always ensuring that samples from $\mathcal{D}_B$ and $\mathcal{D}_C$ are sampled equally.

\section{Experiments}
\label{cegdro:section:experiments}

\paragraph{Datasets and bias-confirming splits.} We benchmark our method on the most popular subpopulation shift setups: Waterbirds \cite{waterbirds}, CelebA \cite{celeba} and CivilComments \cite{civilcomments}. The Waterbirds dataset features a \emph{waterbird} and a \emph{landbird} class strongly correlated in the training set with their respective \emph{water backgrounds} and \emph{land backgrounds}. In order to properly compare ourselves with the environment-aware IRM and GroupDRO approaches, as our method is not concerned with environment discovery, but rather with the post environment discovery optimization procedure, we use the ground-truth annotations and denote \emph{waterbirds on water backgrounds} and \emph{landbirds on landbackgrounds} as bias-confirming and the rest as bias-conflicting. CelebA features two classes, \emph{blond} and \emph{non-blond hair}, with the spurious attributes revolving around \emph{male} and \emph{female} features. For CelebA, $\mathcal{D}_B$ is comprised of \emph{blonde females} and \emph{non-blonde males}. CivilComments features \emph{offensive} and \emph{non-offensive} comments, with spuriousness revolving around references to the following demographics: \emph{(male, female, LGBTQ, Christian, Muslim, other religions, Black, and White)}. We use the standard splits provided by the WILDS benchmark \cite{wilds}. Since the ground-truth annotations provide no clear separation between bias-confirming and bias-conflicting samples, we denote the samples misclassified by $M^\prime$ as bias-conflicting.

\paragraph{Experimental setup.} To ensure fair comparison and reproducibility, we follow the exact setup of Yang et al.~\cite{change_is_hard}. A ResNet50 model pretrained on ImageNet is employed for image datasets and a pretrained BERT model for text datasets. For all benchmarks, we use a curriculum rate $R$ of $0.2$, and we iterate over the samples available at each stage $E_s=8$ times. We train the models for up to $30000$ steps in all cases, except for the Waterbirds dataset, where $1300$ steps is enough to ensure convergence. As with GroupDRO and IRM, our method is only concerned with post environment discovery optimization, thus, ground-truth environments are used in evaluating all methods presented in Table \ref{cegdro:tab:results}. We follow the exact procedure proposed by Yang et al.~\cite{change_is_hard} with regards to optimizing hyperparameters; for each scenario, we perform $16$ runs with different random seeds to determine the best hyperparameter configuration and then report the performance of the model across three runs using the best configuration. Worst group accuracy is used as a selection criteria across all instances.

\begin{table}[t]
\caption{Result on the most popular subpopulation shift datasets. Our proposed method achieves state-of-the-art results across all benchmarks while reducing the variation across runs.}
\label{cegdro:tab:results}
\centering
\setlength{\tabcolsep}{5pt}
\vspace{1mm}
\begin{tabular}{l  c c  c c   c c   c c}
    \toprule
    & \multicolumn{2}{c}{\textbf{Waterbirds}} & \multicolumn{2}{c}{\textbf{CelebA}} & \multicolumn{2}{c}{\textbf{CivilComments}}\\
    & Average & \textbf{Worst Gr.} & Average & \textbf{Worst Gr.} & Average & \textbf{Worst Gr.}\\
    \midrule
    ERM & $84.1${\scriptsize$\pm1.7$} & $69.1${\scriptsize$\pm4.7$} & 95.1{\scriptsize$\pm0.2$} & $62.6${\scriptsize$\pm1.5$} & $85.4${\scriptsize$\pm0.2$} & $63.7${\scriptsize$\pm1.1$} \\
    IRM & $88.4${\scriptsize$\pm0.1$} & $74.5${\scriptsize$\pm1.5$} & $94.7${\scriptsize$\pm0.8$} & $63.0${\scriptsize$\pm2.5$} & 85.5{\scriptsize$\pm0.0$} & $63.2${\scriptsize$\pm0.8$} \\
    GroupDRO & 88.8{\scriptsize$\pm1.8$} & 78.6{\scriptsize$\pm1.0$} & $91.4${\scriptsize$\pm0.6$} & $89.0${\scriptsize$\pm0.7$} & $81.8${\scriptsize$\pm0.6$} & $70.6${\scriptsize$\pm1.2$} \\
    GroupDRO + SC & 67.8{\scriptsize$\pm8.7$} & 49.4{\scriptsize$\pm4.5$} & 86.7{\scriptsize$\pm0.0$} & 0.0{\scriptsize$\pm0.0$} & 82.0{\scriptsize$\pm0.0$} & 0.0{\scriptsize$\pm0.0$} \\
    CeGDRO - EF (ours) & 90.0{\scriptsize$\pm0.4$} & 84.3{\scriptsize$\pm0.7$} & 91.7{\scriptsize$\pm0.2$} & 89.4{\scriptsize$\pm0.7$} & 78.7{\scriptsize$\pm0.5$} & 71.4{\scriptsize$\pm0.8$} \\
    \textbf{CeGDRO (ours)} & 90.3{\scriptsize$\pm0.2$} & \textbf{84.8{\scriptsize$\pm0.6$}} & 91.9{\scriptsize$\pm0.1$} & \textbf{89.8{\scriptsize$\pm0.3$}} & 80.4{\scriptsize$\pm0.2$} & \textbf{73.5{\scriptsize$\pm0.2$}} \\
    \bottomrule
\end{tabular}
\end{table}
    
    
    
    
    
    
\paragraph{Results.} We benchmark our method against the state-of-the-art post environment-discovery optimization protocols: IRM \cite{irm} and GroupDRO \cite{groupdro}. We showcase results in Table \ref{cegdro:tab:results}. We report, for each dataset, the average performance and standard deviation across three runs, measured both in terms of the domain-standard target optimization objective, the worst group accuracy, as defined in Eq.~\eqref{cegdro:eq:objective}, as well as in terms of the average accuracy. In addition to the state-of-the-art results, we also provide as references, the performance levels of: (i) the standard Empirical Risk Minimiazation (ERM) approach, (ii) a Standard Curriculum learning approach, feeding the easiest samples first and gradually proceeding to harder data points, empowered by GroupDRO (GroupDRO + SC), and (iii) a modified version of CeGDRO for which we set the bias-confirming samples to follow an easy-first (CeGDRO - EF) line up. Our proposed Curriculum-enhanced GroupDRO (CeGDRO) approach outperforms the state-of-the-art methods across all scenarios. For example, CeGDRO surpasses GroupDRO by $6.2\%$ on Waterbirds, $0.8\%$ on CelebA, and $2.9\%$ on CivilComments, while improving training stability, reducing the standard deviation of models across multiple runs on all benchmarks.

\section{Conclusion}
We propose Curriculum-enhanced GroupDRO (CeGDRO) as an optimization protocol for bias prevention in subpopulation shift setups. We redesign the curriculum approach for the task at hand, prioritizing the hardest bias-confirming samples and the easiest bias-conflicting samples first, initializing the model weights in an unbiased vantage point in the hypothesis space which sabotages easy convergence towards biased hypotheses during the final optimization stage. To the best of our knowledge, we are the first to adapt and propose a curriculum learning approach for this domain. We benchmark our approach against the most popular subpopulation shift data set, showing an improvement upon the state-of-the-art results across all benchmarks, while improving model stability. We aim to expand upon this work in the future, creating a context for a general bias-prevention curriculum, applicable to all general settings, regardless of the optimization protocol employed.


\small
\bibliography{neurips_2024}
\bibliographystyle{abbrvnat}

\end{document}